\pdfoutput=1

\documentclass[11pt]{article}

\usepackage[preprint]{acl}

\usepackage{times}
\usepackage{latexsym}

\usepackage[T1]{fontenc}

\usepackage[utf8]{inputenc}

\usepackage{microtype}

\usepackage{inconsolata}

\usepackage[dvipsnames]{xcolor}
\usepackage{graphicx}
\usepackage[most]{tcolorbox}

\usepackage{amsmath}
\usepackage{bm}
\usepackage{bbm}
\usepackage{amssymb}
\usepackage{amsthm}
\usepackage{booktabs}
\usepackage{subcaption}
\usepackage[ruled]{algorithm2e}

\usepackage{tikz}
\usepackage{pgfplots}
\pgfplotsset{compat=1.18}
\usetikzlibrary{math,calc,positioning,intersections,fit,matrix,datavisualization,patterns,patterns.meta,decorations.pathreplacing,decorations.markings,arrows.meta,shapes.geometric,pgfplots.groupplots,shapes.misc}
\tikzset{>=latex}

\pgfdeclarelayer{background}
\pgfsetlayers{background,main}

\usepackage{colortbl}
\newcolumntype{a}{>{\columncolor{Gray!20}}c}

\def\eqref#1{equation~\ref{#1}}

\def\1{\bm{1}}

\def\rx{{\textnormal{x}}}

\def\rvx{{\mathbf{x}}}

\def\vzero{{\bm{0}}}
\def\vone{{\bm{1}}}

\def\vc{{\bm{c}}}
\def\vd{{\bm{d}}}

\def\vp{{\bm{p}}}
\def\vq{{\bm{q}}}

\def\vs{{\bm{s}}}

\def\vv{{\bm{v}}}
\def\vw{{\bm{w}}}
\def\vx{{\bm{x}}}

\def\evd{{d}}

\def\evp{{p}}

\def\evw{{w}}
\def\evx{{x}}

\DeclareMathAlphabet{\mathsfit}{\encodingdefault}{\sfdefault}{m}{sl}
\SetMathAlphabet{\mathsfit}{bold}{\encodingdefault}{\sfdefault}{bx}{n}

\newcommand{\R}{\mathbb{R}}

\DeclareMathOperator*{\argmax}{arg\,max}
\DeclareMathOperator*{\argmin}{arg\,min}

\def\vpi{{\bm{\pi}}}
\def\vphi{{\bm{\phi}}}
\usepackage{stmaryrd}
\renewcommand{\P}{\mathbb{P}}
\newcommand{\TV}{D_{\mathrm{TV}}}

\usepackage{mdframed}
\definecolor{theoremcolor}{rgb}{0.95, 0.95, 0.95}
\mdfsetup{
    backgroundcolor=theoremcolor,
    linewidth=0cm
}
\theoremstyle{definition}
\newmdtheoremenv{lemma}{Lemma}
\newmdtheoremenv{theorem}{Theorem}
\newmdtheoremenv{definition}{Definition}

\title{%
\textsc{Kad}: A Framework for Proxy-based Test-time Alignment\\with Knapsack Approximation Deferral}

\author{Ayoub Hammal \and Pierre Zweigenbaum \\
   \\
  \texttt{\{ayoub.hammal,pz\}@lisn.fr} \\\And
  Caio Corro \\
   \\
  \texttt{caio.corro@irisa.fr} \\}

\author{
    Ayoub Hammal\textsuperscript{1} \qquad Pierre Zweigenbaum\textsuperscript{1} \qquad Caio Corro\textsuperscript{2} 
    \\
    \textsuperscript{1}Université Paris-Saclay, CNRS, LISN
    \\
    \textsuperscript{2}INSA Rennes, IRISA, CNRS, Université de Rennes
    \\
    \texttt{\{ayoub.hammal,pz\}@lisn.fr} \qquad\texttt{caio.corro@irisa.fr}
}

\begin{document}
\maketitle
\begin{abstract}
Several previous works concluded that the largest part of generation capabilities of large language models (LLM) are learned (early) during pre-training.
However, LLMs still require further alignment to adhere to downstream task requirements and stylistic preferences, among other desired properties.
As LLMs continue to scale in terms of size, the computational cost of alignment procedures increase prohibitively.
In this work, we propose a novel approach to circumvent these costs via proxy-based test-time alignment,
\emph{i.e.}\ using guidance from a small aligned model.
Our approach can be described as a token-specific cascading method,
where the token-specific deferral rule is reduced to 0-1 knapsack problem.
In this setting, we derive primal and dual approximations of the optimal deferral decision.
We experimentally show the benefits of our method  both in task performance and speculative decoding speed.
\end{abstract}

\section{Introduction}
\label{sec:introduction}

Large language models' (LLM) alignment\footnote{The term ``alignment'' has several different meanings in the literature. In this work, we call alignment the supervised training process of a LLM following the (\textasciitilde{}unsupervised) pre-training phase.} is employed for reshaping the pre-trained model's output distribution so that it adheres to expected (human) preferences, formatting or instructions, and more generally downstream task requirements \citep{antropic2022helpfulharmless, kumar2025llmposttrainingdeepdive, ouyang2022followinstructions}.
A variety of fine-tuning approaches are employed with different computational requirements and overheads, such as reinforcement learning from human feedback \citep[RLHF,][]{ziegler2020finetuninglanguagemodelshuman, rafailov2023dpo, ethayarajh2024kto}.
The alignment phase follows an already costly pre-training phase,
and alignment costs scale prohibitively with model sizes,
meaning that tailoring LLMs for diverse real-world scenarios is challenging.
For example, the RLHF stage of Tulu~3 scales from 520 H100 GPU hours for the 8B model to 2\,880 hours for the 70B model and to 11\,776 hours for the 405B model, following an exponential trend \citep{lambert2025tulu}

To avoid paying the high cost of fine-tuning a large model, previous works have proposed to adapt the pre-trained model's output distribution at test-time \cite{welleck2024metagen},
using external guidance either from a reward signal \cite{deng2023rewardaugmenteddec,troshin2025arm},
implicitly extracting the reward from a small aligned model \cite{liu2024tuningbyproxy,mitchell2024emulator} or using a mixture of a large unaligned model and a small aligned model \cite{lu2023policyadapters,fei2025nudging}.
As such, when novel alignment specifications are needed, one only need to either train a (simple) reward or align a small model, drastically reducing alignment time, costs and even lowering the hardware requirement.

In this work, we follow the \emph{proxy-based alignment} methodology where an unaligned large model distribution is ``distorted'' at test-time using a small aligned model.
One of the motivations of test-time alignment is that previous analysis has shown that most of a LLM's generation capabilities is acquired during pre-training \cite{zhou2023lima},
making the following steps mostly intervene on stylistic and transitional tokens positions \citep{chang2024how, hu2024towards}.
Following this observation, \citet{fei2025nudging} proposed to use the large model's confidence as an alignment indicator:
if the unaligned large model distribution has a low-entropy, the next token is sampled from it,
otherwise the generation is deferred to a smaller aligned model that assists in ``nudging'' the response.

Our main contribution is a novel framework for proxy-based alignment.
Instead of fully deferring the decision to the small model when the large one has a high entropy,
we build a mixture distribution of the two models.
Our intuition is as follows:
we should defer the decision to the small model for low-probability tokens.
We model this deferral rule as a \emph{binary knapsack} problem \cite{martello1990knapsackproblems},
where one may fix a maximum mass (budget) of probability whose decision can be deferred to the small model.
As the binary knapsack problem is known to be intractable \cite{karp1972}, we propose dual and primal approximations of the optimal deferral condition.
Overall, this framework allows us to derive several deferral rules (depending, for example, on the loss function used to measure the quality of generations), including the nudging model of \citet{fei2025nudging}.
Our approach, while adding more degrees of freedom to the deferral rule, allows to focus deferral on specific uncertain tokens, leveraging richer token-level information rather than a single global confidence score.
We furthermore show experimentally that token-specific deferral decision is faster than distribution-level decision in comparable speculative decoding setups.

Our contributions can be summarized as follows:
(1)~we introduce \textsc{Kad} (knapsack approximation deferral), a framework for proxy-based alignment;
(2)~we derive several dual and primal approximate solutions and provide theoretical guarantees on the regret and error-rate achieved by them while linking our propositions with previous work;
(3)~we experiment with OLMo 2 and Qwen 3, showing the performance improvement brought by our approach compared to several baselines;
and (4)~we show our method leads to increased generation speed when combined with speculative decoding.

Our implementation is publicly available.\footnote{\url{https://github.com/ayoubhammal/knapsack-approximation-deferral}}

\paragraph{Notations.}
We write Iverson brackets as $\llbracket \cdot \rrbracket$, taking the value 1 if the inner condition is true and 0 otherwise.
For any $k \in \mathbb N_{\geq 1}$, $\triangle(k)$ is the $(k-1)$-dimensional simplex.
To simplify notations, we denote the token vocabulary as $V = \{1, \dots, |V|\}$.

An autoregressive language model is a conditional distribution of the next token given the previously generated ones.
Let $p$ be such a language model, we can then write the next token distribution as $p(\rx_{t} | \rvx_{1:t-1} = \vx_{1:t-1})$.
To simplify notations, we will drop the dependence on history $\rvx_{1:t-1}$ and write the model's conditional probability distribution as a vector $\vp \in \triangle(|V|)$ defined as follows:
\[
\evp_v = p(\rx_{t} = v | \rvx_{1:t-1} = \vx_{1:t-1}),
\]
\emph{i.e.}\ $\evp_v$ is the probability of token $v$ given history $\vx_{1:t-1}$.
More generally, all variables and functions presented are assumed to be context-dependent unless stated otherwise.

\section{Motivations}

The term \emph{base}\footnote{Also called \emph{pre-trained} language model.} language model (LM) refers to an LM  that has not undergone specific instruction or preference tuning via supervised fine-tuning, reinforcement learning from human feedback (RLHF), or any other related methods.
A model resulting from the latter procedures is called an \textit{aligned} LM.

For the purpose of this paper, we will consider models of different sizes and at different training stages.
We denote $\vp$ and $\vp^*$ a large LM in its base and aligned variants, respectively.
Similarly, $\vq$ and $\vq^*$ refer to small (base and aligned) LMs,
where ``small'' means that generation with these models is significantly faster than with the large ones.
Note that $\vp, \vp^*$, $\vq$ and $\vq^*$ are used to denote both language models and associated conditional distribution vectors.
The only extra assumption we introduce is that the large and small LMs share the same tokenization vocabulary.

\subsection{Background}
\label{sec:background}

In order to avoid high alignment costs,
there is an interest in test-time alignment, where a large base LM is guided to achieve desired outputs directly during generation, without extra tuning steps.
\emph{Proxy tuning} is one such approach, where a small language model is aligned and is used to guide the large one \cite{liu2024tuningbyproxy,mitchell2024emulator}.

Interestingly, several authors hypothesized that the alignment training phase mainly impacts style, discourse structure, and other superficial features, but that most of the actual ``knowledge'' is acquired during the base model training \citep{zhou2023lima, chang2024how, hu2024towards}.
\citet{lin2024theunlockingspell} showed that the alignment procedure changes the base distribution only for a few output positions.
In particular, transitional and structural positions are shifted toward more stylistic tokens such as those showing agreement, \emph{e.g.}\ ``\texttt{of course}'', or positivity, \emph{e.g.}\ ``\texttt{great question}'', conveying less information but helping set the tone of the response.

\citet{fei2025nudging} further show that the base model's certainty is tightly related to the agreement between the base and aligned models' distributions:
the top predictions of both models tend to agree more when the base model's top probability is higher.
From this observation, they propose \textit{nudging} the base model's generation when its certainty falls below a specific threshold $\lambda \in \R_{\geq 0}$, that is, sampling from $q^\ast$ only for positions where $p$ is uncertain.
In other words, the sampling procedure for the next token is:\footnote{In \citet{fei2025nudging}, the nudging step generates the next word until the following space instead of a single token.}
\begin{itemize}
    \item if $\max_v p_v \geq \lambda$, sample from $p$;
    \item otherwise, sample from $q^\ast$.
\end{itemize}

\subsection{Broader Perspective}
\label{sec:broader_perspective}

Nudging can be interpreted as a specific case of \emph{model cascades}, an ensembling technique that relies on a sequence of classifiers:
each classifier in the sequence can choose to either return the prediction or defer the prediction to the next model in the sequence \citep{narasimhan2025fastercascades, varshney2022modelcascading, wang2018idkcascadesfastdeep, li2021cascadebert}.

First, notice that we can rewrite the nudging procedure as sampling from a mixture distribution $\vphi \in \triangle(|V|)$ defined as follows:
\begin{align}\label{eq:nudging}
    \phi_v = \begin{cases}
        q^\ast_v       \quad & \text{if } \max_w p_w < \lambda, \\
        p_v  \quad & \text{otherwise}.
    \end{cases}
\end{align}
The condition $\max_w p_w < \lambda$ is the position-dependent \emph{deferral rule},
whose value indicates the \emph{deferral decision}, that is when the first model in the cascade $\vp$ must defer the decision to the (single) next model $\vq^*$.

From this perspective, we see that nudging as proposed by \citet{fei2025nudging} is a special case of model cascades with a sequence of 2 classifiers, where the second model $\vq^*$ is invoked when the first $\vp$ is uncertain, a condition which is implemented using the so-called Chow's rule \cite{chow1970errorreject,jitkrittum2023confidence,gupta2024language}.
As such, an obvious question is whether we can use different deferral rules.
\section{Knapsack Approximation Deferral}
\label{sec:token_level_mixture}

As described in the previous section, the nudging technique for proxy-based alignment can be interpreted as a specific case of cascading.
Based on this observation, we first introduce a framework that allows to derive token-specific mixtures for test-time alignment based on a loss function that measures the base model's uncertainty, and a maximum deferral budget.
Then, we propose two approximation methods of the optimal deferral decision.

\subsection{Formalization}

We seek to build a mixture distribution $\vpi$ using a token-specific deferral decision $\vd \in \{0, 1\}^{|V|}$ \citep{narasimhan2025fastercascades}:
\begin{align}\label{eq:pi}
\pi_v = p_v(1 - \evd_v) + q^\ast_v \alpha
\end{align}
where $\alpha$ is a normalization term that ensures the distribution $\pi$ is well-defined:
\[
\alpha = \sum_{w \in V} p_w \evd_w.
\]
The deferral decision $\vd$ allows to defer decision to $\vq$ independently for each token, and as such generalizes the nudging procedure described in Sec.~\ref{sec:broader_perspective}.

We are now left with defining rules used to compute the deferral decision vector $\vd$.
As explained in Sec.~\ref{sec:background}, alignment seems to only shift the probability of a few tokens, especially when the model is \emph{unsure} about an outcome.
To measure the uncertainty about a specific output, we rely on a loss function $\ell: \triangle(k) \times V \to \R_{\geq 0}$.
We define the \emph{risk} \cite{shalev2014understandingml} associated with $\vd$ as:
\[
r(\vd)
=
    \sum_{v \in V} \P_v \ell(\vp, v)(1 - \evd_v),
\]
where $\P_v = \P(\rx_{t} = v | \rvx_{1:t-1} = \vx_{1:t-1})$ is the (unknown) ground-truth target distribution.

Notice that naive risk minimization leads to a trivial deferral decision as $r(\vone) = 0$,
\emph{i.e.}\ ignoring the base model $\vp$ and deferring the outcome to $\vq^*$ only.
Instead, we assume a maximum deferral budget $b > 0$, that is the maximum probability mass that can be ignored in $\vp$.
We obtain the following constrained risk minimization problem:
\begin{align*}
    (P1)\quad\min_{\vd \in \{0, 1\}^{|V|}}~
    r(\vd)
    \quad
    \text{s.t.}
    \quad
    \sum_{v \in V} \P_v \evd_v \leq b.
\end{align*}
We denote $\widehat \vd$ an optimal solution of $(P1)$.

\begin{lemma}
    Computing the optimal deferral rule $\widehat \vd$ is NP-Hard.
\end{lemma}

The lemma can be proved by observing that (P1) is equivalent to the 0-1 knapsack optimization problem, which is known to be NP-Hard \citep{cocchiani2022knapsackproblems}.
Although exact dynamic programming algorithms are known for solving this problem, their worst-case complexity is exponential in the vocabulary size \cite{martello1990knapsackproblems,pisinger1997dpknsapsack,martello1999dpknapsack}.
Therefore, we will instead rely on fast dual and primal approximations, which are guaranteed to define bounds on the optimal risk.

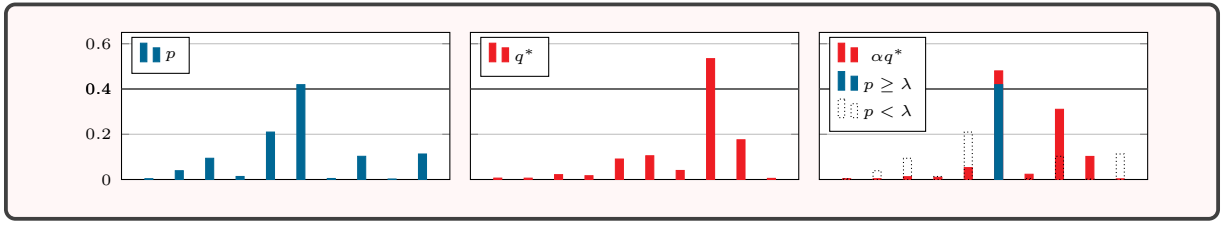
\begin{figure*}[t!]
    \begin{tcolorbox}[colback=red!3!white]
    \centering
        \scalebox{1.0}{
\begin{tikzpicture}
    \begin{axis}[
        ybar,
        tiny,
        name=plot1,
        yminorgrids,
        ymajorgrids,
        bar width=3.0pt,
        bar shift auto,
        x=0.4cm,
        y=3.0cm,
        xtick=\empty,
        legend pos=north west,
        legend columns=1,
        extra y ticks={0.4},
        extra y tick style={grid style={black}},
        ymin=0,
        ymax=0.65,
        axis background/.style={fill=white},
    ]
        \addplot+[MidnightBlue] table [x=x, y=y] {figures/summary_p.dat};
        \addlegendentry{$p$};
    \end{axis}
    \begin{axis}[
        ybar,
        tiny,
        at=(plot1.right of north east),
        anchor=left of north west,
        name=plot2,
        yminorgrids,
        ymajorgrids,
        bar width=3.0pt,
        bar shift auto,
        x=0.4cm,
        y=3.0cm,
        xtick=\empty,
        legend pos=north west,
        legend columns=1,
        extra y ticks={0.4},
        extra y tick style={grid style={black}},
        yticklabel=\empty,
        ymin=0,
        ymax=0.65,
        axis background/.style={fill=white},
    ]
        \addplot+[Red] table [x=x, y=y] {figures/summary_q.dat};
        \addlegendentry{$q^\ast$};
    \end{axis}
    \begin{axis}[
        ybar,
        tiny,
        at=(plot2.right of north east),
        anchor=left of north west,
        name=plot3,
        yminorgrids,
        ymajorgrids,
        bar width=3.0pt,
        bar shift=0pt,
        x=0.4cm,
        y=3.0cm,
        xtick=\empty,
        legend pos=north west,
        legend columns=1,
        extra y ticks={0.4},
        extra y tick style={grid style={black}},
        yticklabel=\empty,
        ymin=0,
        ymax=0.65,
        axis background/.style={fill=white},
    ]
        \addplot+[Red] table[x=x, y=y] {figures/summary_pi.dat};
        \addlegendentry{$\alpha q^\ast$};
        \addplot+[MidnightBlue] table [x=x, y=y] {figures/summary_p_d0.dat};
        \addlegendentry{$p \geq \lambda$};
        \addplot[densely dotted] table [x=x, y=y] {figures/summary_p_d1.dat};
        \addlegendentry{$p < \lambda$}
    \end{axis}
\end{tikzpicture}
}
    \end{tcolorbox}
    \caption{%
    Illustration of token-specific cascading distribution $\vpi^{<\lambda}$ with $\lambda = 0.4$.
    The left and center plots show the probability mass function (PMF) of $\vp$ and $\vq^*$, respectively.
    The right plot shows the PMF of $\vpi^{<\lambda}$,
    where the blue parts shows the mass coming from $\vp$ and the red parts the one coming from $\vq^*$ (including rescaling by $\alpha$).
    Dotted bars show that mass from $\vp$ that was rejected by the deferral decision.
    We can observe the most probable token of $\vp$ and the one of $\vq^*$ both have high probabilities in the resulting mixture.
    }
    \label{fig:distribution_level}
\end{figure*}

\subsection{Dual Approximation}

In this section, we introduce a first approximation based on duality.
This approximation does not rely on the ground-truth distribution $\P$.

By relaxing the budget constraint, we build the following Lagrangian \cite{lemarechal2001lr}:
\begin{align*}
    L(\vd, \mu)
    &= r(\vd) + \mu \left (~ \sum_{v \in V} \P_v \evd_v - b ~\right ),
\end{align*}
where $\mu \in \R_{\geq 0}$ is called a Lagrangian multiplier or dual variable.
In other words, the hard constraint has been replaced by a penalty in the objective.
Our dual approximation is simply defined as a dual solution  for a given multiplier $\mu$:
\[
\widehat \vd^-\left(\mu\right) \in \argmin_{\vd \in \{0, 1\}^{|V|}} L(\vd, \mu).
\]
Note that $\widehat\vd^-\left(\mu\right) \in \R^{|V|}$ may not be primal feasible,
\emph{i.e.}\ it may not satisfy the budget constraint.

\begin{lemma}\label{lemma:dual_bound}
    For any dual variable $\mu \geq 0$, the deferral decision vector $\widehat\vd^-(\mu)$ is defined as:
    \begin{align*}
        \forall v \in V: \left[ \widehat\vd^-({\mu})\right]_v &= \llbracket \ell(\vp, v) > \mu \rrbracket.
    \end{align*}
    where the right-hand side condition is called the deferral condition.
    Moreover, the following bound on the true risk holds:
    \[
        L\left(\widehat\vd^-(\mu), \mu\right) \leq r(\widehat \vd).
    \]
\end{lemma}
Proof of Lemma~\ref{lemma:dual_bound} can be found in Appendix~\ref{app:proof_dual_bound}.
For the negative log-likelihood (or logistic) loss,
we can rewrite the dual approximation as follows:
\begin{align*}
    \left[ \widehat\vd^-({\mu})\right]_v
    &= \llbracket - \log p_v > \mu \rrbracket
    = \llbracket p_v < \exp(- \mu) \rrbracket.
\end{align*}
For any loss of the form $\ell(\vp, v) = f(p_v)$
where $f$ is continuous and strictly decreasing,
we have:
\begin{equation}
    \label{eq:dual_approx}
    \left[ \widehat\vd^-({\mu})\right]_v
    = \llbracket p_v < f^{-1}(\mu) \rrbracket.
\end{equation}
To simplify notation, we will denote $\vpi^{<\lambda}$ the distribution given by such decision rule:
\[
\pi^{<\lambda}_v =
p_v \times \llbracket p_v \geq \lambda \rrbracket + q^*_v \alpha,
\]
where $\lambda = f^{-1}(\mu)$.
Figure~\ref{fig:distribution_level} illustrates such a mixture.

Further examples with different losses are presented in Table \ref{tab:dual_def_rules} and their derivations in App.\ \ref{app:proof_dual_def_rules}.

\begin{table}[t!]
    \small
    \centering
    \begin{tabular}{l@{\hskip 0.3em}c@{\hskip 0.3em}c}
    \toprule
        \textbf{Loss} & $\ell(\vp, v)$ & $\left[ \widehat\vd^-(\mu) \right]_v$
    \\
    \midrule
        Logistic
        & $- \log p_v$
        & $\llbracket p_v < \exp(-\mu) \rrbracket$
        \\
        Squared
        & $(1 - p_v)^2$
        & $\llbracket p_v < 1 - \sqrt{\mu} \rrbracket$
        \\
        Perceptron
        & $- p_v + \max_w p_w$
        & $\llbracket p_v < \max_w p_w - \mu \rrbracket$
        \\
        $0$-$1$
        & $\llbracket v \notin \argmax_w p_w \rrbracket$
        & $\llbracket p_v < \max_w p_w \rrbracket$
        \\
        \midrule
        Min-ent.
        & $- \log \max_w p_w$
        & $\llbracket \max_w p_w < \exp(-\mu) \rrbracket$
        \\
        \bottomrule
    \end{tabular}
    \caption{Different losses and their associated deferral rules in the dual approximation scenario. For the $0$-$1$ loss, we assume that $\mu \in (0, 1)$.}
    \label{tab:dual_def_rules}
\end{table}

\paragraph{Chow's rule.}
We now show that our framework generalizes the nudging method of \citet{fei2025nudging}.
Consider the min-entropy loss function \cite{renner2004minent} defined as follows:
\[
    \ell(\vp, v) = - \log~\max_{w \in V}~p_w\,.
\]
Note that this loss function is unsupervised,
in the sense that the loss value does not depend on the gold token $v$.
With this loss, the deferral decision is:
\[
    \left[ \widehat\vd^-({\mu})\right]_v
    = \llbracket \max_{w \in V} p_w < \exp(- \mu) \rrbracket,
\]
which is Chow's rule used for nudging, which we recover here as a dual approximation.

\subsection{Primal Approximation}

The proposed dual approximation does not depend on $\P$, but requires setting a hyperparameter.
In this section, we derive a method to build a primal feasible solution using the Lagrangian dual problem.
As computing the solution requires access to the ground-truth distribution,
we rely on a plug-in approximation of $\P$ in practice.

From Lemma~\ref{lemma:dual_bound}, we know that for any $\mu \in \R_{\geq 0}$, the Lagrangian gives a lower bound to the optimal risk.
As such, we seek the dual variable value $\widehat\mu$ that maximizes this bound:
\begin{align*}
    \widehat\mu \in \argmax_{\mu \geq 0} L(\mu),
\end{align*}
where $L(\mu) = \min_{\vd \in \{0,1\}^{|V|}} L(\vd, \mu)$.

\begin{lemma}\label{lemma:optimal_dual_bound}
Wlog, we assume the vocabulary is sorted in increasing loss order,
that is:
\[
\forall v, w \in V:\quad v < w \implies \ell(\vp, v) < \ell(\vp, w).
\]
Let $\widehat k$ be defined such that the following condition holds:
\begin{align}
    \sum_{v=\widehat k+1}^{|V|} \P_v \leq b < \sum_{v=\widehat k}^{|V|} \P_v. \label{eq:critical_condition}
\end{align}
Then, we have
\(
\widehat\mu = \ell(\vp, \widehat k).
\)
We call $\widehat k$ the critical element index.
\end{lemma}
The proof can be found in Appendix~\ref{app:optimal_dual_bound}.

The primal approximation is then defined as:
\[
\widehat \vd^+ = \widehat \vd^- (\widehat \mu).
\]
As $\widehat \evd^+_v = \llbracket \ell(\vp, v) > \ell(\vp, \widehat k) \rrbracket$, and assuming that the vocabulary is sorted in increasing loss order, we can see that the following inequality holds:
\[
    \sum_{v \in V} \widehat \evd^+_v \P_v = \sum_{v = \widehat k + 1}^{|V|} \P_v \leq b.
\]
Thus $\widehat \vd^+$ is primal feasible,
and the following bound holds:
\[
    r(\widehat\vd) \leq r(\widehat\vd^+).
\]

\paragraph{Plugin approximation.}
Unfortunately, computing the critical element index $\widehat k$ requires access to the ground-truth data distribution $\P$, which is unknown.
Therefore, we follow the methodology of \citet{narasimhan2025fastercascades} and use a plugin approximation of $\P$, that is we rely on $\vp$ as an approximation of $\P$.
More precisely, we compute an approximation $\widetilde \mu$ of the optimal Lagrangian multiplier using an approximation of the  critical index element $\widetilde k$:
\[
\widetilde\mu = \ell(\vp, \widetilde k)\quad\text{s.t.}\quad\sum_{v=\widetilde k+1}^{|V|} p_v \leq b < \sum_{v=\widetilde k}^{|V|} p_v.
\]
The primal approximation $\widetilde \vd^+ = \widehat \vd^- (\widetilde \mu)$ of the deferral rule is defined as follows:\footnote{Pseudo-code is given in Appendix~\ref{app:primal_approx_sol}.
}
\[
\forall v\in V: \widetilde \evd^+_v = \llbracket \ell(\vp, v) > \widetilde \mu\rrbracket.
\]

\subsection{Theoretical Analysis}

In this section, we show some theoretical properties of primal approximations.

First, it is interesting to note that the previously considered decreasing loss functions all produce the same primal approximation as per Lemma \ref{lemma:order_inverting_losses}.
Therefore, although there is an interest of evaluating different losses for the dual approximation, the primal one always lead to the same deferral rule for a wide class of losses (Lemma~\ref{lemma:order_inverting_losses}).
Second, we show that the quality of the approximation using $\vp$ as a plugin estimator depends on the total variation distance between $\vp$ and $\P$ (Lemma~\ref{lemma:regret_bound_primal}).
\begin{lemma}
    \label{lemma:order_inverting_losses}
    A loss function $\ell$ is \emph{order inverting} if, $\forall v, w \in V$, we have:
    \[
        \evp_v < \evp_w
        \iff
        \ell(\vp, w) < \ell(\vp, v).
    \]
    Let  $\ell_1$ and $\ell_2$ be two order inverting losses.
    Then, $\ell_1$ and $\ell_2$ produce the same critical element index $\widehat k$ and the same approximation $\widetilde k$.
    Moreover, they produce the same deferral decision $\widehat\vd^+$, and the same approximation $\widetilde\vd^+$.
\end{lemma}

\begin{lemma}
\label{lemma:regret_bound_primal}
The following bound on the risk holds: 
\begin{align*}
    |r(\widehat \vd^+) - r(\widetilde \vd^+)| \leq 
    \ell(\vp, u) \left(\hspace{-0.3em}
        \begin{array}{l}
            \TV(\P, \vp) \\
            + \sum_{v = u}^{l} \P_v
        \end{array}\hspace{-0.3em}
    \right),
\end{align*}
where $l = \min(\widehat k, \widetilde k) + 1$ and $u = \max(\widehat k, \widetilde k)$.
Moreover, we have:
    \[
        r(\widehat \vd^+) - r(\widehat \vd) \leq \ell(\vp, \widehat k) \P_{\widehat k}\,.
    \]
\end{lemma}

Proofs are given in Appendices~\ref{app:proof_order_inverting_losses} and~\ref{app:proof_regret_bound_primal}.

\section{Speculative Decoding}
\label{sec:speculative}

In the nudging approach of \citet{fei2025nudging},
they only need to forward in the large model $\vp$ when its certainty is high, and calls to $\vq^*$ are limited to a few tokens for which the decision is deferred to it, see Equation~(\ref{eq:nudging}).
On the contrary, our mixture distribution defined in Equation~(\ref{eq:pi}) relies on both $\vp$ and $\vq^*$ as soon as there is a single element of the deferral decision equal to 1.
Hence, a naive use of our approach will lead to a slower generation.

In this section, we propose to accelerate generation using \emph{speculative decoding} \cite{leviathan2023fastinference,chen2023speculativedec}, a two step sampling process.
First, in the drafting step, a \emph{draft model} is used to samples a draft of $\gamma \in \mathbb{Z}_{>1}$ tokens.
Then, in the verification step, token are sequentially accepted using a \emph{target model}, until one token is rejected.
The resulting sampling process is guaranteed to match the target model distribution.
In practice, the draft model is assumed to be small and fast, whereas the target model is large and slow.
The forward pass in the target model can be efficiently parallelized over all tokens in the draft instead of generating tokens autoregressively, which leads to experimental speed improvement.
We refer readers to \cite{leviathan2023fastinference} for further details.

In our case, the target distribution is the mixture $\vpi$, and the proposal distribution is the aligned model $\vq^*$.
The probability $a(\vpi, \vq^*) \in [0, 1]$ of accepting the next token in the draft is equal to:
\[
    a(\vpi, \vq^*) = 1 - \TV(\vpi, \vq^*).
\]
As such, a lower divergence between the target and the draft models means a higher acceptance probability, and therefore a faster speculative procedure.

\begin{lemma}\label{lemma:tv}
    The following bound on $\TV(\vpi, \vq^*)$ holds:
    \[
        \TV(\vpi^{<\lambda}, \vq^\ast) \leq \TV(\vphi, \vq^\ast) + \alpha (1 - \beta),
    \]
    where $\beta = \sum_v \evd_v q^\ast_v$.

    Moreover, if the following condition are satisfied:
    \begin{enumerate}
        \item $\max_{v \in V} p_v \geq \lambda$\,,
        \item $\forall v: p_v < \lambda \implies p_v \leq \alpha q^\ast_v$\,,
    \end{enumerate}
    then, the following bound holds:
    \[
    \TV(\vpi^{<\lambda}, \vq^\ast) \leq \TV(\vphi, \vq^\ast).
    \]
\end{lemma}

The proof can be found in Appendix~\ref{app:tv_comparison}.
Lemma~\ref{lemma:tv} shows that even if our mixture distribution $\vpi^{<\lambda}$ is more expressive than the nudging distribution $\vphi$,
the speculative decoding process will only have a slightly lower acceptance rate, as $\alpha (1\!-\!\beta) \ll 1$.
The second part of theorem even show that in many cases, the acceptance will be at least as good, and even better.
We will confirm these results experimentally in Section~\ref{sec:experiment}.
\section{Related work}
\label{sec:related_work}
\paragraph{Test-time alignment.}
To avoid the high cost of LLM alignment via parameter fine-tuning, different previous work considered to leverage a reward signal to steer the generation of the base model.
Unlike our approach, these methods require a specifically trained reward model to evaluate partial sequences to guide the decoding step from the base model \citep{khanov2024args, deng2023rewardaugmenteddec,troshin2025arm}.
\citet{mitchell2024emulator} and \citet{liu2024tuningbyproxy} proposed extracting the partial reward from a smaller aligned model, but their approach requires both the base and aligned checkpoints of the smaller model.
Alternatively, \citet{faria2025sampledontsearchrethinking} proposed a Metropolis-Hastings sampling approach that can use a more standard reward model defined on complete outputs.

Our test-time alignment method is more straightforward as it only requires the aligned small model and isolates its contribution to the final mixture distribution to the weak components of $\vp$.
Additionally, it does not require sampling several responses for each prediction.

\paragraph{Model cascading.}
Model cascading is a popular technique used in computer vision \citep{wang2018idkcascadesfastdeep} and natural language processing \citep{varshney2022modelcascading, li2021cascadebert}, with an emphasis on reducing generation cost by evaluating larger models only when necessary.
By design, cascading can only defer to larger models using information about the the smaller model outputs.
For text generation, \citet{narasimhan2025fastercascades} proposed to mix cascading and speculative decoding to build deferral rules based on larger model outputs while ensuring fast generation.
Our work differ in its goal, that is proxy-based test-time alignment of a large base model, which leads to a different definition of the deferral rule and resulting output distributions.

\paragraph{Knapsack formulation.}
\citet{nishikawa2014learning} and \citet{tonglet2023seer} proposed reductions to the 0-1 knapsack problem for summarization and in-context examples selection, respectively.

\setlength{\tabcolsep}{2.5pt}
\begin{table*}[!ht]
    \small
    \centering
    \begin{tabular}{l ccc ccc a c ccc ccc a}
        \toprule
        & \textbf{GMS8K} & \textbf{MATH} & \textbf{SVAMP} & \textbf{ARC} & \textbf{CSQA} & \textbf{TQA} & \textbf{Avg.} && \textbf{GMS8K} & \textbf{MATH} & \textbf{SVAMP} & \textbf{ARC} & \textbf{CSQA} & \textbf{TQA} & \textbf{Avg.} \\
        \cmidrule(lr){2-8}  \cmidrule(lr){10-16}
        & \multicolumn{7}{c}{OLMo 2 (1B and 13B)} & & \multicolumn{7}{c}{Qwen 3 (1.7B and 14B)} \\
        \midrule
        \multicolumn{16}{l}{\textbf{Building blocks}}\\
        \midrule
        $\vp$                      & $54.5$ & $09.4$ & $57.6$ & $29.6$ & $19.4$ & $29.3$ & $33.3$ && $75.5$ & $51.8$ & $80.0$ & $86.6$ & $76.9$ & $57.8$ & $71.4$ \\
        $\vq^\ast$                 & $62.5$ & $16.4$ & $70.3$ & $43.8$ & $48.4$ & $22.3$ & $43.9$ && $75.3$ & $53.0$ & $86.6$ & $82.9$ & $68.7$ & $43.2$ & $68.2$ \\
        
        \midrule
        \multicolumn{16}{l}{\textbf{Nudging \cite{fei2025nudging}}}\\
        \midrule
        
        $\lambda\!=\!0.3$ & $60.3$ & $13.8$ & $67.3$ & $53.7$ & $33.4$ & $38.5$ & $44.5$ && $78.1$ & $52.8$ & $85.0$ & $88.8$ & $78.5$ & $57.0$ & $73.3$ \\
        $\lambda\!=\!0.4$ & $61.9$ & $18.8$ & $66.6$ & $57.8$ & $46.3$ & $38.5$ & $48.3$ && $79.5$ & $57.8$ & $87.3$ & $89.6$ & $79.3$ & $\mathbf{60.7}$ & $75.7$ \\

        \midrule
        \multicolumn{16}{l}{\textbf{Our work: dual approximation}}\\
        \midrule
        \multicolumn{2}{l}{$\hookrightarrow$ 0-1 loss}
        &&&&&&&&&&&&&&
        \\[0.1cm]
        
        $\mu\!\in\!(0, 1)$              & $68.8$ & $\mathbf{24.6}$ & $74.3$ & $58.9$ & $48.7$ & $38.3$ & $52.2$ && $81.7$ & $59.2$ & $\mathbf{89.6}$ & $\mathbf{92.1}$ & $\mathbf{82.0}$ & $58.1$ & $77.1$ \\[0.2cm]
        
        \multicolumn{3}{l}{$\hookrightarrow$ perceptron loss}
        &&&&&&&&&&&&&
        \\[0.1cm]
        $\mu = 0.05$     & $68.2$ & $22.0$ & $74.3$ & $55.8$ & $46.1$ & $37.6$ & $50.6$ && $\mathbf{82.4}$ & $59.6$ & $\mathbf{89.6}$ & $91.4$ & $81.4$ & $\underline{60.0}$ & $\mathbf{77.4}$ \\
        $\mu = 0.1$      & $67.7$ & $21.4$ & $73.6$ & $57.3$ & $44.6$ & $39.2$ & $50.6$ && $\underline{82.0}$ & $\underline{60.0}$ & $\underline{89.0}$ & $\mathbf{92.1}$ & $\underline{81.9}$ & $58.6$ & $\underline{77.2}$ \\[0.2cm]
        
        \multicolumn{3}{l}{$\hookrightarrow$ $\vpi^{<\lambda}$ with $\lambda = f^{-1}(\mu)$}
        &&&&&&&&&&&&&
        \\[0.1cm]
        $\lambda = 0.3$  & $\underline{69.5}$ & $22.8$ & $\mathbf{76.6}$ & $\underline{61.6}$ & $\underline{52.3}$ & $\mathbf{42.8}$ & $\underline{54.2}$ && $81.3$ & $59.6$ & $\underline{89.0}$ & $91.4$ & $81.3$ & $59.6$ & $77.0$ \\
        $\lambda = 0.4$  & $\mathbf{72.3}$ & $23.4$ & $\underline{75.3}$ & $\mathbf{61.9}$ & $\mathbf{55.6}$ & $\underline{40.2}$ & $\mathbf{54.7}$ && $81.7$ & $\mathbf{60.6}$ & $87.3$ & $\underline{91.5}$ & $80.7$ & $58.3$ & $76.6$ \\
        
        \midrule
        \multicolumn{16}{l}{\textbf{Our work: primal approximation}}\\
        \midrule
        
        $b\!=\!0.6$        & $65.5$ & $21.0$ & $74.3$ & $54.9$ & $42.8$ & $38.1$ & $49.4$ && $81.6$ & $59.8$ & $87.6$ & $91.3$ & $81.4$ & $58.3$ & $76.6$ \\
        $b\!=\!0.9$        & $68.3$ & $\underline{23.8}$ & $70.6$ & $59.0$ & $49.2$ & $38.6$ & $51.5$ && $81.6$ & $\mathbf{60.6}$ & $88.6$ & $90.7$ & $81.5$ & $59.7$ & $77.1$ \\

        \midrule
        \multicolumn{16}{l}{\textbf{For reference}}\\
        \midrule
        $\vp^\ast$                 & $84.3$ & $39.6$ & $87.6$ & $82.5$ & $76.9$ & $53.6$ & $70.7$ && $82.4$ & $64.0$ & $88.3$ & $93.8$ & $83.1$ & $70.1$ & $80.2$ \\
        $\vq$                      & $14.4$ & $01.0$ & $17.6$ & $19.6$ & $15.8$ & $18.8$ & $14.5$ && $21.9$ & $12.2$ & $25.3$ & $47.1$ & $29.5$ & $15.0$ & $25.1$ \\
        Imp.\ reward                    & $58.4$ & $18.2$ & $73.0$ & $63.3$ & $55.8$ & $30.1$ & $49.8$ && $80.7$ & $60.6$ & $89.0$ & $88.9$ & $78.1$ & $55.2$ & $75.4$ \\

        \bottomrule
    \end{tabular}
    \caption{Model's accuracy. Best results on each column are in bold, second best are underlined.}
    \label{tab:accuracy}
\end{table*}

\section{Experiment}
\label{sec:experiment}

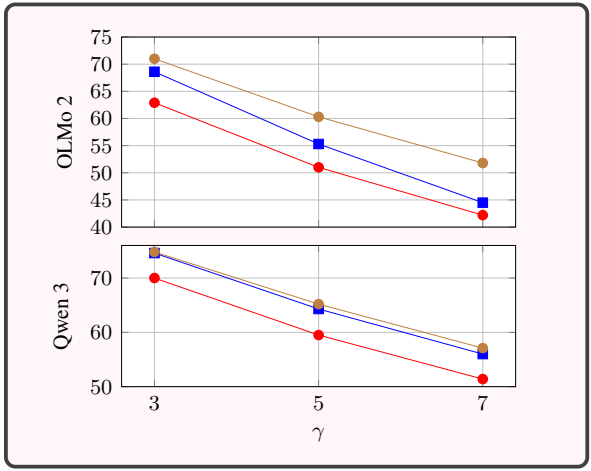
\begin{figure}
    \centering
    \begin{tcolorbox}[boxsep=0pt, colback=red!3!white]
    \scalebox{0.9}{
\begin{tikzpicture}
\begin{axis}[
    footnotesize,
    ymax=75,
    ymin=40,
    name=olmo,
    x=1.2cm,
    y=0.08cm,
    ylabel={OLMo 2},
    grid=both,
    axis background/.style={fill=white},
    xtick={3, 5, 7},
    xticklabel=\empty,
]
    \addplot+[red] coordinates {
        (3, 62.9)
        (5, 51.0)
        (7, 42.2)
    };
    \addplot+[blue] coordinates {
        (3, 68.6)
        (5, 55.3)
        (7, 44.5)
    };
    \addplot+[brown] coordinates {
        (3, 71.0)
        (5, 60.3)
        (7, 51.8)
    };
\end{axis}
\begin{axis}[
    footnotesize,
    ymax=76,
    ymin=50,
    at=(olmo.below south east),
    anchor=above north east,
    name=qwen,
    x=1.2cm,
    y=0.08cm,
    xlabel={$\gamma$},
    ylabel={Qwen 3},
    grid=both,
    axis background/.style={fill=white},
    xtick={3, 5, 7},
]
    \addplot+[red] coordinates {
        (3, 70.0)
        (5, 59.5)
        (7, 51.4)
    };
    \addplot+[blue] coordinates {
        (3, 74.6)
        (5, 64.3)
        (7, 56.0)
    };
    \addplot+[brown] coordinates {
        (3, 74.8)
        (5, 65.2)
        (7, 57.1)
    };
\end{axis}
\end{tikzpicture}
}
    \end{tcolorbox}
    \caption{Empirical acceptance rate per model in speculative generation schema. In brown the dual approximation $\vpi^{<\lambda}$ with $\lambda = 0.4$, in blue the primal approximation with $b = 0.9$ and in red Nudging with $\lambda = 0.4$.}
    \label{fig:acceptance_rate}
\end{figure}

\paragraph{Datasets.}
We evaluate our method on three \textit{math reasoning} datasets, GSM8K \citep{cobbe2021trainingverifierssolvemath}, MATH500 \citep{lightman2023letsverifystepstep} and SVAMP \citep{patel2021svamp}, and three \textit{commonsense reasoning} datasets, ARC-Challenge \citep{allenai2018arc}, CommonsenseQA \citep{talmor2019commonsenseqa} and TruthfulQA \citep{lin2022truthfulqa}.
We use standard prompting and evaluation strategies for all datasets.
Examples are given in Appendix~\ref{app:dataset_examples}.

\paragraph{Models.}
We use two LLM famillies for experiments.
First, OLMo 2 \citep{olmo20252olmo2furious}, which is a fully open source model family.
Second, Qwen 3 \citep{yang2025qwen3technicalreport}, which is relatively closed source.
We select 2 contrasting sizes of each family on which we carry out our experiments.
The choice of the model families is based on the available model sizes and checkpoints.

Unlike \citet{fei2025nudging}, we do not employ greedy decoding.
Instead, we sample from the unmodified distributions with temperature $0.7$ as recommended for each model to isolate the benefits of our method.

All the experiments where run on single Nvidia A100-80GB GPUs.
The evaluation of each dataset was parallelized on 8-GPU nodes.

\paragraph{Test-time methods.}
For our approach, we evaluate both with dual and primal approximations.
For the dual approximation case, we test 3 loss functions.
For the primal approximation case, we shown that all (non-degenerated) losses lead to the same deferral rule, therefore will only test with the negative log-likelihood.

We compare our approach against two baselines:
(1) nudging \citep{fei2025nudging},
and (2) the implicit reward approach introduced by \citet{mitchell2024emulator} and \citet{liu2024tuningbyproxy}.
The implicit reward builds a mixture in which the probability of token $v \in V$ is proportional to $p_v\frac{q^*_v}{q_v}$.
Note this approach requires access to both $\vq^*$ and $\vq$, whereas nudging and ours only require access to $\vq^*$

\paragraph{Hyperparameters.}
For our mixture $\vpi^{<\lambda}$, we follow \citet{fei2025nudging} and set $\lambda \in \{0.3, 0.4\}$.
We do not explore further specific tuning for our token-specific deferral rule, showing that it can serve as a plug-in replacement for the distribution-level deferral rule with no further hyperparameter search.
As for the other deferral rules, we performed a quick hyperparameter scan on a small data subset and selected promising hyperparameter values, that is $b \in \{0.6, 0.9\}$ and $\mu \in \{0.05, 0.1\}$.

\subsection{Performance evaluation}
The main experimental results are summarized in Table~\ref{tab:accuracy}.
As expected, an aligned model always performs better than its base model, and a larger base (resp.\ aligned) model always performs better than its smaller base (resp.\ aligned) model.  With OLMo~2, $\vq^\ast$ is nearly always better than the larger base model $\vp$, whereas this only happens in two out of six datasets for Qwen~3.  For a given model, scores vary widely across datasets, \emph{e.g.}\ from 9.4 (MATH500) to 57.6 (SVAMP) for the base OLMo~2.  Therefore, it is convenient to focus on average scores over the six datasets.

All test-time alignment methods in the table consistently reach an average performance above that of the base $\vp$ and $\vq$ and of the small aligned $\vq^\ast$ model: in other words, they all play the expected role.
While our dual approximation allows to recovers Chow's rule used in the nudging method of \citet{fei2025nudging}, our other proposed generalizations approximations obtain average scores that are always above that of the nudging method.
They are also always above that of the implicit reward method except once (primal approximation with OLMo~2 and $b = 0.6$ results in lower accuracy than implicit reward).

Overall, our approach obtains the highest average accuracy for OLMo~2 and also the highest average over OLMo~2 and Qwen~3.

For Qwen~3, each primal approximation obtains at least one average score within the small range $[77.0, 77.4]$.  This plateau effect might be related to the relatively small distance between $\vp$ (71.4) and $\vp^\ast$ (80.2), where the base model $\vp$ already performs fairly high, and to the fact that the small aligned model $\vq^\ast$ does not outperform the larger base model $\vp$, hence possibly making it more difficult to provide relevant guidance to it.
In contrast, for OLMo-2, the best average score of each primal approximation varies within the larger range $[50.6, 54.7]$.  This might be linked to the much larger gap between $\vp$ (33.3) and $\vp^\ast$ (70.7) and to the fact that the small aligned model $\vq^\ast$ performs better than the larger base model $\vp$.

\subsection{Speed evaluation}

Figure~\ref{fig:acceptance_rate} shows the empirically calculated acceptance rates for $\vphi$ and $\vpi^{<\lambda}$ across different speculative window sizes $\gamma \in \{3, 5, 7\}$.
We can see in this figure that $\vpi^{<\lambda}$ always achieve higher acceptance rates than $\vphi$ across the 3 window sizes, with our dual approximation achieving slightly higher acceptance rates than the primal approximation.

This higher acceptance rate for $\vpi^{<\lambda}$ translates into higher throughput in speculative generation mode as shown in Table~\ref{tab:throughput} in Appendix~\ref{app:generation_speed}.

\section{Further Analysis}
\label{app:analysis}

\paragraph{Conditional distribution entropy.}
The Shannon entropy is defined as follows:
\[
H[\vp] = -\sum_{v \in V} p_v \log p_v.
\]
Entropy is a measure of uncertainty, \emph{i.e.}\ high entropy means high uncertainty.
In the following we consider entropies of conditional distributions,
where we condition the distribution on sampling element that are deferred in $\vpi^{<\lambda}$, for different values of $\lambda$, which we write as follows:
\begin{align*}
    &\left[\langle \vp, \vd\rangle^{-1} \vp \odot \vd\right]_v
    \\
    &\quad= p\left(
        \rx_{t} = v
        \middle|
        \begin{array}{l}
        \rx_t \in \{ v \in V | d_v = 1\}, \\\rvx_{1:t-1} = \vx_{1:t-1}
        \end{array}
    \right),
\end{align*}
where $\odot$ is the element-wise multiplication,
$\langle \cdot, \cdot \rangle$ is the dot product.
We define a similar conditional distribution for $\vp^*$.

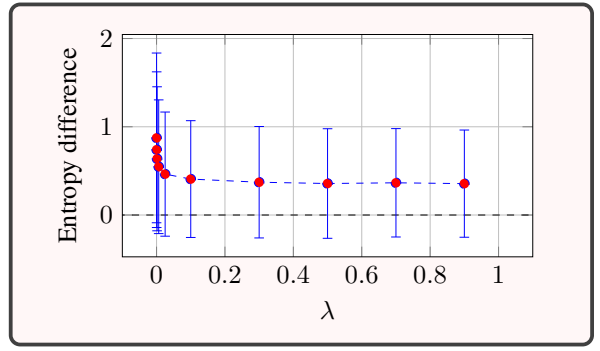
\begin{figure}
    \centering
    \begin{tcolorbox}[boxsep=0pt, colback=red!3!white]
    \scalebox{0.9}{
\begin{tikzpicture}
\begin{axis}[
    xmin=-0.1,
    xmax=1.1,
    y=1.3cm,
    x=5.0cm,
    xlabel={$\lambda$},
    ylabel style={align=center},
    ylabel={Entropy difference},
    grid=both,
    axis background/.style={fill=white},
]
    \addplot [domain=-1:2, samples=2, dashed, black] {0};
    \addplot+[
        color=blue,
        dashed,              
        mark=*,
        mark options={fill=red},
        error bars/.cd,
            y dir=both,
            y explicit,
            error bar style={solid}
    ] table [x=x, y=y, y error=ey] {figures/diff_entropy.dat};
\end{axis}
\end{tikzpicture}
}
    \end{tcolorbox}
    \caption{We plot the mean entropy difference $H\left[ \langle \vp, \vd\rangle^{-1} \vp \odot \vd \right] - H\left[ \langle \vp^*, \vd\rangle^{-1} \vp^* \odot \vd \right]$ for different values of $\lambda$.
    We report mean differences evaluated on OLMo-2 1B and 13B on MATH500.}
    \label{fig:diff_entropy}
\end{figure}

When the following inequality holds:
\[
H\left[ \langle \vp, \vd\rangle^{-1} \vp \odot \vd \right]
\geq
H\left[ \langle \vp^*, \vd\rangle^{-1} \vp^* \odot \vd \right],
\]
the low probability elements of $\vp$ has a higher uncertainty than the same elements of $\vp^*$.
To check if this is true empirically,
we plot in Figure~\ref{fig:diff_entropy} the difference of entropy between the conditional distributions based on $\vp$ and $\vp^*$, for different values of $\lambda$.
We can see the entropy of the conditional distribution based on $\vp$ is much higher than that of its aligned version $\vp^\ast$.
This motivates the need to mask those parts of $\vp$ and replace them with $\vq^\ast$.

\paragraph{$\vp$-selection proportion.}
Table~\ref{tab:p_selection_prop} shows the number of tokens whose mass from the large base model $p$ is kept in the mixture, \emph{i.e.}\ the number of non-deferred tokens.
We show statistics for $\pi^{< \lambda}$ with $\lambda = 0.4$,
and for primal approximation with a budget of $\{0.6, 0.9\}$.
We can see that most of the time, the mass of a single token is kept.
For the primal approximation case, setting a budget of $b=0.9$ is even almost equivalent to a deferral decision that only keeps the top-1 token.

\begin{table}[]
    \centering
    \small
    \begin{tabular}{l@{\hspace{3em}}c@{\hspace{1em}}c@{\hspace{1em}}c@{\hspace{1em}}c}
        \toprule
        & \multicolumn{4}{c}{$\|\vone - \vd\|_1$}\\
        \cmidrule(lr){2-5}
         & 0 & 1 & 2 & 3\\
        \midrule
        \multicolumn{5}{l}{\textbf{Dual approximation}}\\
        \midrule
        \multicolumn{5}{l}{$\hookrightarrow$ $\vpi^{<\lambda}$ with $\lambda = f^{-1}(\mu)$}\\[0.1em]
        $\lambda = 0.4$
        & $9.5\%$ & $88.6\%$ & $1.9\%$ & $0\%$
        \\
        \midrule
        \multicolumn{5}{l}{\textbf{Primal approximation}}\\
        \midrule
        $b\!=\!0.6$ & $0.0\%$ & $90.7\%$ & $6.3\%$ & $1.5\%$ \\
        $b\!=\!0.9$ & $0.0\%$ & $99.9\%$ & $0.1\%$ & $0\%$ \\
        \bottomrule
    \end{tabular}
    \caption{Number of selected tokens from the large base model $p$ for different deferral rules.
    Evaluated on OLMo-2 1B and 13b on MATH500.}
    \label{tab:p_selection_prop}
\end{table}

\paragraph{Budget selection.}
Setting a high budget for the primal approximation does not mean that it will necessarily be fully met, as the remaining unfilled budget depends on the probability (in the large base model $p$) of the critical element.
An example is shown in Figure~\ref{fig:critical_element_illustration}.

\begin{figure}
    \centering
    \begin{tcolorbox}[boxsep=0pt, colback=red!3!white]
    \begin{tikzpicture}[font=\small]

\def\B{6.7}
\def\b{0.6 * \B}
\def\barh{0.4}

\def\basea{0}
\def\baseb{-1.5}

\def\paa{0.1 * \B}
\def\pab{0.2 * \B}
\def\pac{0.27 * \B}
\def\pad{0.43 * \B}

\def\pba{0.1 * \B}
\def\pbb{0.2 * \B}
\def\pbc{0.32 * \B}
\def\pbd{0.38 * \B}

\node[anchor=west] at (0, \basea+\barh+0.3) {\textbf{Small unfilled budget}};
\draw[thick, black, fill=SkyBlue!30] (0,\basea) rectangle (\paa,\basea+\barh) node[midway, black] {$\evp_1$};
\draw[thick, black, fill=SkyBlue!30] (\paa,\basea) rectangle (\paa+\pab,\basea+\barh) node[midway, black] {$\evp_2$};
\draw[thick, black, fill=SkyBlue!30] (\paa+\pab,\basea) rectangle (\paa+\pab+\pac,\basea+\barh) node[midway, black] {$\evp_3$};
\draw[thick, black, fill=BrickRed!30] (\paa+\pab+\pac,\basea) rectangle (\paa+\pab+\pac+\pad,\basea+\barh) node[midway, black] {$\evp_4$};

\draw[pattern=north west lines, pattern color=black!70] (\paa+\pab+\pac,\basea) rectangle (\b,\basea+\barh);

\draw[dashed, thick, black] (\b, \basea-0.25) -- (\b, \basea+\barh+0.25);
\node at (\b, \basea+\barh+0.4) {Budget};

\node[anchor=west] at (0,\baseb+\barh+0.3) {\textbf{Large unfilled budget}};

\draw[thick, black, fill=SkyBlue!30] (0,\baseb) rectangle (\pba,\baseb+\barh) node[midway, black] {$\evp_1$};
\draw[thick, black, fill=SkyBlue!30] (\pba,\baseb) rectangle (\pba+\pbb,\baseb+\barh) node[midway, black] {$\evp_2$};
\draw[thick, black, fill=BrickRed!30] (\pba+\pbb,\baseb) rectangle (\pba+\pbb+\pbc,\baseb+\barh) node[midway, black] {$\evp_3$};
\draw[thick, black, fill=BrickRed!30] (\pba+\pbb+\pbc,\baseb) rectangle (\pba+\pbb+\pbc+\pbd,\baseb+\barh) node[midway, black] {$\evp_4$};

\draw[pattern=north west lines, pattern color=black!70] (\pba+\pbb,\baseb) rectangle (\b,\baseb+\barh);

\draw[dashed, thick, black] (\b, \baseb-0.25) -- (\b, \baseb+\barh+0.25);

\draw[thick, pattern=north west lines, pattern color=black!70] (0,\baseb-1) rectangle (0.5,\baseb-1+\barh);
\node[anchor=north west, align=left] at (0.5+0.1,\baseb-1+\barh) {Unfilled budget};

\draw[thick, fill=SkyBlue!30] (3,\baseb-1) rectangle (3.5,\baseb-1+\barh);
\node[anchor=north west, align=left] at (3.5+0.1,\baseb-1+\barh) {$\evd_i = 1$};

\draw[thick, fill=BrickRed!30] (5,\baseb-1) rectangle (5.5,\baseb-1+\barh);
\node[anchor=north west, align=left] at (5.5+0.1,\baseb-1+\barh) {$\evd_i = 0$};

\end{tikzpicture}
    \end{tcolorbox}
    \caption{Illustration of the effect of the critical element probability on the filled budget. A small difference in $p_3$ can have a big impact on the unfilled budget.}
    \label{fig:critical_element_illustration}
\end{figure}
\section{Conclusion}
In this paper, we introduce a novel framework for test-time alignment based on a 0-1 knapsack problem.
This framework allows to derive several dual and primal approximations,
that all take the form of a token-specific deferral rules similar to cascading \cite{narasimhan2025fastercascades}.
We proposed different instances of these deferral rules, generalizing existing work, and evaluated experimentally their performance against strong baselines.

\section*{Acknowledgments}

We thank Miguel Couceiro for his comments and suggestions.

This work is supported by the SEMIAMOR (CE23-2023-0005) and InExtenso (ANR-23-IAS1-0004) project grants given by the French National Research Agency (ANR).
This work was granted access to the HPC resources of IDRIS under the allocation 2024-AD011015801 made by GENCI.

\section*{Limitations}
While our approach cuts training costs by a considerable margin, it draws its guidance from a smaller less capable model and it slightly underperforms the larger aligned model $\vp^\ast$.
Although hyperparameters choices such as that of the budget $b$ and the threshold $\lambda$ does not heavily influence the mixture performance, they still need to be tuned individually for each model and dataset on a separate validation set.
We however show in this paper that a fixed hyperparameter choice can safely be carried between models and datasets with marginal performance degradation.

\bibliography{custom}

\appendix

\section{Proof of Lemma \ref{lemma:dual_bound}}
\label{app:proof_dual_bound}

\begin{proof}
The Lagrangian can be rewritten as follows:
\begin{align*}
    L(\vd, \mu)
    =& r(\vd) + \mu \left(~ \sum_{i \in V} \P_i \evd_i - b ~\right)
    \\
    =& \sum_{v \in V} \evd_v \P_v (\mu - \ell(p, v)) \\\
    &+ \sum_{v \in V} \P_v \ell(p, v) - \mu b,
\end{align*}
where only the first sum depends on $\vd$.
Therefore, computing the dual approximation can be rewritten as:
\[
\widehat{\vd}^-(\mu) \in \argmin_{\vd \in \{0, 1\}^{|V|}} \sum_{v \in V} \evd_v \P_v (\mu - \ell(p, v)),
\]
where, for all $v \in V$, $\P_v \geq 0$ and $\ell(p, v) \geq 0$ by definition.
Therefore, to minimize the sum in the objective, we must set $\evd_v = 1$ if $\mu - \ell(p, v) < 0$, and 0 otherwise.
As such, we can write:
\begin{gather*}
    \left[ \widehat\vd^-({\mu})\right]_v = \llbracket \ell(p, v) > \mu \rrbracket
    \\
    \implies
    \\
    \widehat\vd^-({\mu}) \in \argmin_{\vd \in \{0, 1\}^{|V|}} L(\vd, \mu).
\end{gather*}
Note that if we use the condition:
\[
\left[ \widehat\vd^-({\mu})\right]_v = \llbracket \ell(p, v) \geq \mu \rrbracket,
\]
although it does not change the objective value, it increases the probability of violating the budget constraint.

The bound on the true risk follows standard linear optimization properties \cite{lemarechal2001lr}.
As $\widehat\vd$ satisfy the budget constraint, we have $\sum_{v \in [V]} \P_v \widehat\evd_v - b \leq 0$, therefore:
\begin{align*}
    r(\widehat\vd)
    &\geq r(\widehat\vd) + \mu \left(~ \sum_{v \in V} \P_v \widehat\evd_v - b ~\right).
    \\
\intertext{We derive a lower bound on this term by minimizing on the deferral decision, which leads to the desired bound:}
    &\geq \min_{\vd \in \{0, 1\}^{|V|}} L(\vd, \mu)
    \\
    &= L(\widehat \vd^-(\mu), \mu).
\end{align*}
\end{proof}

\section{Deferral Rules for Dual Approximation}
\label{app:proof_dual_def_rules}

The deferral rule for dual approximation has the following form:
\[
    \left[ \widehat \vd^-(\mu)\right]_v = \llbracket\ell(\vp, v) > \mu\rrbracket.
\]
We derive the particular form of this rule for the losses that are summarized in Table~\ref{tab:dual_def_rules}.

\paragraph{Negative log-likelihood (logistic).}
Let $\ell$ be defined as:
\[
\ell(\vp, v) = - \log p_v.
\]
Then we can rewrite the deferral condition as:
\begin{align*}
    \ell(\vp, v) &> \mu,
    \\
    - \log p_v &> \mu,
    \\
    p_v &< \exp(-\mu),
\end{align*}
which shows in this case the rule defer to $\vq^\ast$ for low probability tokens.

\paragraph{Squared.}
Let $\ell$ be defined as:
\[
\ell(\vp, v) = (1 - p_v)^2
\]
Then we can rewrite the deferral condition as:
\begin{align*}
    \ell(\vp, v) &> \mu,
    \\
    (1 - p_v)^2 &> \mu,
    \\
    1 - p_v &> \sqrt\mu,
    \\
    p_v &< 1 - \sqrt\mu.
\end{align*}
which shows in this case the rule defer to $\vq^\ast$ for low probability tokens.

Note the similarity with the condition of the logistic loss.
Given $\mu, \mu' \in \R_{\geq 0}$, and the two conditions:
\begin{align}
    p_v &< \exp(-\mu) \label{eq:cond_eq1} \\
    \text{and}\quad p_v &< 1 - \sqrt{\mu'}. \label{eq:cond_eq2} 
\end{align}
If the following condition holds:
\begin{align}
    \mu &= - \log\left(1 - \sqrt{\mu'}\right),
    \\
\intertext{or equivalently:}
    \mu' &= (1 - \exp(-\mu))^2,
\end{align}
then the two deferral rules are equivalent.

\paragraph{Perceptron.}
Let $\ell$ be defined as:
\[
\ell(\vp, v) = - p_v + \max_{w\in V} p_w
\]
Then we can rewrite the deferral condition as:
\begin{align*}
    \ell(\vp, v) &> \mu,
    \\
    - p_v + \max_{w\in V} p_w &> \mu,
    \\
    p_v &< \max_{w\in V} p_w - \mu,
\end{align*}
which shows in this case the rule defer to $\vq^\ast$ for tokens whose probability is lower than the max token probability minus a margin given by the dual variable $\mu$.

\paragraph{Zero-one.}
Let $\ell$ be defined as:
\[
\ell(\vp, v) = \llbracket v \notin \argmax_{w \in V} p_w \rrbracket
\]
and since $\mu \in (0, 1)$ and $\ell(\vp, v) \in \{0, 1\}$, $\left [ \widehat \vd^-(\mu) \right ]_v = 1$ if and only if $\ell(\vp, v) = 1$ which is the case when $v \notin \argmax_{w\in V} p_w$ or equivalently when $p_v < \max_{w\in V} p_w$.

\section{Proof of Lemma~\ref{lemma:optimal_dual_bound}}
\label{app:optimal_dual_bound}

We give here a proof of Lemma~\ref{lemma:optimal_dual_bound} using first order optimality conditions. A more visual explanation for a similar problem can be found in \citep{Maculan01111983}.

\begin{proof}
First, note that $L$ is a concave function since it is a pointwise minimum over a family of affine functions \cite[][Sec.~3.2.3]{boyd2004convex}:
\begin{align*}
    L(\mu) = \min_{\vd \in \{0, 1\}^{|V|}}~&
    \sum_{v \in V} \P_v \ell(\vp, v)(1 - \evd_v)
    \\
    &\quad + \mu \left ( \sum_{v \in V} \P_v \evd_v - b \right ).
\end{align*}
Per Lemma~\ref{lemma:dual_bound},
for a given dual variable $\mu \in \R_{\geq 0}$,
the optimal minimizer is given by $\widehat \vd^- (\mu)$ defined as:
\[
\left[ \widehat \vd^- (\mu) \right ]_v = \llbracket \ell(\vp, v) > \mu \rrbracket.
\]
We observe that the value of $L$ changes at breakpoints given by $\ell(\vp, v)$, $\forall v \in V$.
The subdifferential set at $\mu \in \R_+$ is given by \cite[][Thm.~3.50]{beck2017firstorder}:
\[
    \partial L(\mu)
    =
    \left[~
        \sum_{v: \ell(\vp, v) > \mu} \P_v - b,
        \sum_{v: \ell(\vp, v) \geq \mu} \P_v - b
    ~\right]\,,
\]
which is the convex hull of derivatives of functions attaining the minimum.

A sufficient condition for $\widehat \mu \in \R_{\geq 0}$ to maximize $L$ is that $0 \in \partial L(\widehat \mu)$.
Assuming that the vocabulary is sorted in increasing loss order and knowing that $\widehat k$ is defined as per condition (\ref{eq:critical_condition}), we have
\[
    \sum_{v = \widehat k}^{|V|} \P_v - b > 0 \quad \text{and} \quad \sum_{v = \widehat k + 1}^{|V|} \P_v - b \leq 0.
\]
Thus, $\widehat \mu = \ell(\vp , \widehat k)$.
\end{proof}


\section{Primal Approximation with the 0-1 Loss}
\label{app:primal_01}

Let $\ell(\vp, v)$ be the 0-1 loss function, that is:
\[
\ell(\vp, v) = \llbracket v \notin \argmax_{w\in V} \evp_w \rrbracket.
\]
As the optimal Lagrangian multiplier has the form:
\begin{align}\label{eq:app_optimal_mu}
\widetilde\mu = \ell(\vp, \widetilde k),
\end{align}
where $k$ is the critical index element,
therefore we know that $\widetilde\mu \in \{0, 1\}$.

If $\widetilde\mu = 1$, then we have:
\[
\left[\widetilde\vd^+\right]_v = \llbracket \ell(\vp, v) > 1 \rrbracket = 0,
\]
as, by definition, the 0-1 loss cannot be strictly greater than 1.
In other word, we defer all the decision to $p$.

If $\widetilde\mu = 0$, then we have:
\[
\left[\widetilde\vd^+\right]_v
= \llbracket \ell(\vp, v) > 0 \rrbracket
= \llbracket v \notin \argmax_{w \in V} \evp_w\rrbracket,
\]
that is, we keep from $\vp$ only the mass of the most probable element.

We are left with finding when we have $\widetilde\mu = 0$.
By Eq.~\eqref{eq:app_optimal_mu} and the definition of the $0$-$1$ loss,
this can only happen if $\widetilde k = 1$.
Remember that the critical index element $\widetilde k$ is defined such that it satisfies the following conditions:
\[
\sum_{v = \widetilde k+1}^{|V|} \evp_v \leq b < \sum_{w = \widetilde k}^{|V|} \evp_v .
\]
Hence, to have $\widetilde k = 1$, we must have $\evp_1 \geq 1 - b$.
We end up with the rule:
\[
\left[\widetilde\vd^+\right]_v
= \llbracket \evp_1 \geq 1 - b
\wedge v \notin \argmax_w \evp_w
\rrbracket.
\]

\section{Proof of Lemma \ref{lemma:order_inverting_losses}}
\label{app:proof_order_inverting_losses}

\begin{proof}
    We give the proof for the primal feasible solution $\widehat\vd^+$, and a similar reasoning holds for the approximate solution $\widetilde\vd^+$.

    According to the hypothesis on $\ell_1$ and $\ell_2$,
    sorting tokens in the increasing order of $\ell_1(\vp, v)$ or $\ell_2(\vp, v)$ result in the same ordering.
    Therefore, they both lead to the same critical element index $\widehat k$, which depends only on $\P$ and the ordering (and not the particular loss values).

    Thus, the optimal Lagrangian multipliers given by the two losses are $\widehat \mu_1 = \ell_1(\vp, \widehat k)$ and $\widehat \mu_2 = \ell_2(\vp, \widehat k)$.
    Since the losses have the same order, it is easy to see that for any $v \in V$,
    we have
    \[
    \llbracket \ell_1(\vp, v) > \widehat \mu_1 \rrbracket
    =
    \llbracket \ell_2(\vp, v) > \widehat \mu_2 \rrbracket,
    \]
    and therefore both losses produces the same primal feasible solution $\widehat\vd^+$.
\end{proof}

\section{Proof of Lemma  \ref{lemma:regret_bound_primal}}
\label{app:proof_regret_bound_primal}
\begin{proof}
Let us consider the quantity 
\[
    |r(\widehat \vd^+) - r(\widetilde \vd^+)|.
\]
Assuming that vocabulary is sorted in increasing loss order, we can see that
\begin{align*}
    |r(\widehat \vd^+)&\ - r(\widetilde \vd^+)|
    \\
    =&\ |r(\widehat \vd^- (\widehat \mu)) - r(\widehat \vd^-(\widetilde \mu))|
    \\
    \leq&\ \sum_v \P_v \ell(\vp, v) \left | [\widehat \vd^-(\widehat \mu)]_v - [\widehat \vd^-(\widetilde \mu)]_v \right |
\intertext{Since $\widehat \vd^- (\widehat \mu)$ and $\widehat \vd^- (\widetilde \mu)$ agree on all positions except $\{\min(\widehat k, \widetilde k) + 1, \dots, \max(\widehat k, \widetilde k)\}$}
    =&\ \sum_{v = \min(\widehat k, \widetilde k) + 1}^{\max(\widehat k, \widetilde k)} \P_v \ell(\vp, v).
\intertext{Let us define $l = \min(\widehat k, \widetilde k) + 1$ and $u = \max(\widehat k, \widetilde k)$, remember that the vocabulary is sorted in increasing loss order:}
    \leq&\ \ell(\vp, u) \sum_{v = l}^{u} \P_v.
\end{align*}
Let us consider the second factor in the left side of the inequality
\begin{align*}
    \sum_{v = l}^{u} \P_v
    &= \frac{1}{2} \sum_{v=1}^{|V|} (\P_v - p_v) + \frac{1}{2} \sum_{v=1}^{|V|} p_v 
    \\
    &\quad + \frac{1}{2} \sum_{v = l}^{u} \P_v - \frac{1}{2} \sum_{v = 1: v \notin [u, l]}^{|V|} \P_v
    \\
    &\leq \TV(\P, p)
    \\
    &\quad + \frac{1}{2} \left ( 1 + \sum_{v = l}^{u} \P_v - \sum_{v = 1: v \notin \{u, \dots l\}}^{|V|} \P_v \right )
    \\
    &= \TV(\P, p) + \sum_{v = l}^{u} \P_v,
\end{align*}
where $\TV$ is the total-variation distance.
Going back to our initial bound
\begin{align*}
    & |r(\widehat \vd^+) - r(\widetilde \vd^+)| \leq 
    \\
    & \quad \ell(\vp, u) \left ( \TV(\P, p) + \sum_{v = l}^{u} \P_v \right ).
\end{align*}
As for the second part of the Lemma, we have 
    \begin{align*}
        L(\widehat \mu) &= L(\widehat \vd^+, \widehat \mu)
        \\
        &= r(\widehat \vd^+) + \widehat \mu \left ( \sum_{i \in V} \P_i \widehat \evd^+_i - b \right )
        \\
        &\leq r(\widehat \vd).
\intertext{From the last inequality, we have}
        r(\widehat \vd^+) - r(\widehat \vd) &\leq \widehat \mu \left ( b - \sum_{i \in V} \P_i \widehat \evd^+_i \right )
        \\
        &\leq \ell(\vp, \widehat k) \P_{\widehat k}.
    \end{align*}
    Since $b < \sum_{i = \widehat k}^{|V|} \P_i \implies b - \sum_{i = \widehat k + 1}^{|V|} \P_i < \P_{\widehat k}$.
\end{proof}

\section{Primal Approximation Algorithm}
\label{app:primal_approx_sol}
Algorithm \ref{alg:critical_element_plugin} shows a simple sorting-based procedure for calculating the primal approximation deferral rule.

\begin{algorithm*}
    \DontPrintSemicolon
    \KwData{Budget $b \in [0, 1]$, distribution $p$.}
    \KwResult{Deferral rule $\vd \in \{0, 1\}^V$.}
    \Begin{
        $\vv \leftarrow - p \log p$
        \tcc*{Values, $p$ is a plug-in estimator for $\P$}\;
        $\vw \leftarrow p$
        \tcc*{Weights, $p$ is a plug-in estimator for $\P$}\;
        $\vc \leftarrow \vv / \vw$ which is equal to $- \log p$
        \tcc*{Densities}\;
        $\vs \leftarrow \operatorname{argsort\_descending}(\vc)$\;

        $\vd \leftarrow \vzero$\;
        $b' \leftarrow 0$\;
        $i \leftarrow 1$\;
        \While{$i \leq V$ and $b' + \evw_{s_i} \leq b$}{
            $\evd_{s_i} \leftarrow 1$\;
            $b' \leftarrow b' + \evw_{s_i}$\;
        }
        \Return $\vd$\;
    }
    \caption{Returns the deferral rule using primal approximation.}
    \label{alg:critical_element_plugin}
\end{algorithm*}

\section{Proof of lemma \ref{lemma:tv}}
\label{app:tv_comparison}

\begin{proof}
    For the first part of the lemma, notice that if $\max_w p_w < \lambda$ then $\forall v \in V, p_v < \lambda$ and $\vd = \vone$ giving $\vphi = \vpi^{<\lambda} = \vq^\ast$ (trivial case).
    On the other hand, if $\max_w p_w \geq \lambda$ then $\vphi = \vp$ and we have:
    \begin{align*}
        \frac{1}{2} \sum_i &\ (1 - \evd_i) |p_i + q^\ast_i \alpha - q^\ast_i|
        \\[0.5cm]
        \leq &\ \frac{1}{2} \sum_i (1 - \evd_i) |p_i - q^\ast_i|
        \\
        &+ \frac{1}{2} \alpha \sum_i (1 - \evd_i) q^\ast_i
        \\[0.5cm]
        \leq &\ \frac{1}{2} \sum_i |p_i - q^\ast_i| - \frac{1}{2} \sum_i \evd_i |p_i - q^\ast_i|
        \\
        &+ \frac{1}{2} \alpha - \frac{1}{2} \alpha \sum_i \evd_i q^\ast_i
        \\[0.5cm]
        = &\ \TV(\vp, \vq^\ast) - \frac{1}{2} \sum_i \evd_i |p_i - q^\ast_i|
        \\
        &+ \frac{1}{2} \alpha - \frac{1}{2} \alpha \sum_i \evd_i q^\ast_i.
    \end{align*}
    Using this result we derive
    {
    \allowdisplaybreaks 
    \begin{align*}
        & \TV(\vpi^{<\lambda}, \vq^\ast)
        = \ \frac{1}{2} \sum_i \evd_i | q^\ast_i - q^\ast_i \alpha |
        \\
        & + (1 - \evd_i) | p_i + q^\ast_i \alpha - q^\ast_i |
        \\[0.5cm]
        \leq &\ \TV(\vp, \vq^\ast) - \frac{1}{2} \sum_i \evd_i |p_i - q^\ast_i|
        \\
        &+ \frac{1}{2} \alpha - \frac{1}{2} \alpha \sum_i \evd_i q^\ast_i 
        \\
        &+ \frac{1}{2} \sum_i \evd_i q^\ast_i - \frac{1}{2} \alpha \sum_i \evd_i q^\ast_i
        \\[0.5cm]
        = &\ \TV(\vp, \vq^\ast) - \frac{1}{2} \sum_i \evd_i |p_i - q^\ast_i|
        \\
        &+ \frac{1}{2} \alpha - \alpha \sum_i \evd_i q^\ast_i + \frac{1}{2} \sum_i \evd_i q^\ast_i
        \\[0.5cm]
        = &\ \TV(\vp, \vq^\ast) - \frac{1}{2} \sum_i \evd_i |p_i - q^\ast_i|
        \\
        &- \alpha \sum_i \evd_i q^\ast_i + \frac{1}{2} \sum_i \evd_i |p_i + q^\ast_i|
        \\[0.5cm]
        \leq &\ - \alpha \sum_i q^\ast_i \evd_i + \sum_i \evd_i p_i + \TV(\vp, \vq^\ast)
        \\[0.5cm]
        = &\ \alpha(1 - \sum_i q^\ast_i \evd_i) + \TV(\vp, \vq^\ast),
    \end{align*}
    }
    which shows that $\TV(\vpi^{<\lambda}, \vq^\ast) \leq \alpha(1 - \sum_i q^\ast_i \evd_i) + \TV(\vphi, \vq^\ast).$

    For the second part of the lemma, remember $\TV(\vp, \vq^\ast) = 1 - \sum_v \min (p_v, q^\ast_v)$.

    We have $\TV(\vpi^{<\lambda}, \vq^\ast) \leq \TV(\vp, \vq^\ast)$ is equivalent to
    $$
        \sum_v \min (\pi^{<\lambda}_v, q^\ast_v) \geq \sum_v \min (p_v, q^\ast_v)
    $$
    which is equivalent to
    \begin{align}
        & \sum_{v \in \mathcal A} \left (~ \min (p_v + \alpha q^\ast_v, q^\ast_v) - \min (p_v, q^\ast_v) ~\right )
        \nonumber \\
        & + \sum_{v \in \mathcal B} \left (~ \alpha q^\ast_v - \min (p_v, q^\ast_v) ~\right ) \geq 0\,,
        \label{eq:inequality_tv}
    \end{align}
    where $\mathcal A = \{v \in V: p_v \geq \lambda\}$ and $\mathcal B = \{v \in V: p_v < \lambda\}$.
    Suppose that the conditions of the lemma hold, then:
    \begin{itemize}
        \item For $v \in \mathcal A$, if $\min(p_v, q^\ast_v) = q^\ast_v$ then $\min(p_v + \alpha q^\ast_v, q^\ast_v) = q^\ast_i$.
        Otherwise if $\min(p_v, q^\ast_v) = p_v$ then $\min(p_v + \alpha q^\ast_v, q^\ast_v) \geq p_v$.
        Hence $\min(p_v + \alpha q^\ast_v, q^\ast_v) - \min(p_v, q^\ast_v) \geq 0$.

        \item For $v \in \mathcal B: p_v \leq \alpha q^\ast_v$ implies $\min (p_v, q^\ast_v) = p_v$ and $\min (\pi_v, q^\ast_v) = \min (\alpha q^\ast_v, q^\ast_v) = \alpha q^\ast_v$. Thus
        $$
            \alpha q^\ast_v - \min (p_v, q^\ast_v) = \alpha q^\ast_v - p_v \geq 0\,.
        $$
    \end{itemize}
    Summing the nonnegative coordinate differences over $\mathcal A$ and $\mathcal B$ gives the left hand side of (\ref{eq:inequality_tv}) is $\geq 0$. Hence $\TV(\vpi^{<\lambda}, \vq^\ast) \leq \TV(\vp, \vq^\ast)$.
\end{proof}

%

\section{Dataset Examples}
\label{app:dataset_examples}

Following previous work, we rely on regular expressions to extract the final response form the LLM generation.
In the case of GSM8K and SVAMP, we extract the last number in the model's response.
In the case of MATH500, we extract the last \texttt{\textbackslash boxed\{\}} number in the model's response.
In the multiple-choice question tasks, we extract the last capital letter from \texttt{A} to \texttt{J} in the response.

The models are evaluated on the question prompts alone, with no additional few-shot examples.
For each task, we use the extracted answer's accuracy to evaluate the model's performance.
We also evaluate the generation throughput as the average number of tokens generated by the model each second.

The following system prompt was used in all tasks:
{
\footnotesize
\begin{verbatim}
Answer the question by walking through
the reasoning step by step.
\end{verbatim}
}

\paragraph{MATH500.}
Here is an example of a question/answer from MATH500 dataset:
{
\footnotesize
\begin{verbatim}
{
  "role": "user",
  "content": "Convert the point $(0,3)$ in
  rectangular coordinates to polar coordinates.
  Enter your answer in the  form $(r,\\theta),$
  where $r > 0$ and $0 \\le \\theta < 2 \\pi.$
  \n\nPresent the answer in LaTex format:
  \\boxed{Your answer}"
},
{
  "role": "assistant",
  "content": "We have that $r = \\sqrt{0^2
  + 3^2} = 3.$ Also, if we draw the line
  connecting the origin and $(0,3),$ this line
  makes an angle of $\\frac{\\pi}{2}$ with the
  positive $x$-axis.\n\n[asy]\nunitsize(0.8 cm);
  \n\ndraw((-0.5,0)--(3.5,0));\ndraw((0,-0.5)
  --(0,3.5)); \ndraw(arc((0,0),3,0,90),red,
  Arrow(6));\n\ndot((0,3), red);\nlabel(\"$(0,3)$
  \", (0,3), W);\ndot((3,0), red); \n[/asy]
  \n\nTherefore, the polar coordinates are
  $\\boxed{\\left( 3, \\frac{\\pi}{2} \\right)}.$"
}
\end{verbatim}
}
The following instruction is appended at the end of the user's query :
{
\footnotesize
\begin{verbatim}
\n\nPresent the answer in LaTex format:
\\boxed{{Your answer}}
\end{verbatim}
}

\paragraph{GSM8K.}
Here is an example of a question/answer from GSM8K dataset:
{
\footnotesize
\begin{verbatim}
{
  "role": "user",
  "content": "Janet\u2019s ducks lay 16 eggs per day.
  She eats three for breakfast every morning and bakes
  muffins for her friends every day with four. She
  sells the remainder at the farmers' market daily
  for $2 per fresh duck egg. How much in dollars does
  she make every day at the farmers' market?"
},
{
  "role": "assistant",
  "content": "How many eggs does Janet sell? ** Janet
  sells 16 - 3 - 4 = <<16-3-4=9>>9 duck eggs a day.
  \nHow much does Janet make at the farmers' market?
  ** She makes 9 * 2 = $<<9*2=18>>18 every day at the
  farmer\u2019s market.
  \n#### 18"
}
\end{verbatim}
}

\paragraph{ARC.}
Here is an example of a question/answer from ARC dataset:
{
\footnotesize
\begin{verbatim}
{
  "role": "user",
  "content": "Choose the correct answer to the
  following multiple-choice question.
  \n\nQuestion: An astronomer observes that a
  planet rotates faster after a meteorite impact.
  Which is the most likely effect of this increase
  in rotation?\n\nA). Planetary density will
  decrease.\nB). Planetary years will become longer.
  \nC). Planetary days will become shorter.
  \nD). Planetary gravity will become stronger.
  \n\nProvide your reasoning about the answer and
  finish your answer with the letter corresponding
  to the correct option (e.g., A, B, C, or D).\n\n"
},
{
  "role": "assistant",
  "content": "\nAnswer: C\n\n"
}
\end{verbatim}
}

The following prefix is prepended to the user's query
{
\footnotesize
\begin{verbatim}
Choose the correct answer to the following
multiple-choice question.\n\n
\end{verbatim}
}
The following suffix is appended to the user's query
{
\footnotesize
\begin{verbatim}
\nProvide your reasoning about the answer and finish
your answer with the letter corresponding to the
correct option (e.g., A, B, C, or D).\n\n
\end{verbatim}
}

\setlength{\tabcolsep}{2.5pt}
\begin{table*}[!ht]
    \small
    \centering
    \begin{tabular}{l ccc ccc a c ccc ccc a}
        \toprule
        & \textbf{GMS8K} & \textbf{MATH} & \textbf{SVAMP} & \textbf{ARC} & \textbf{CSQA} & \textbf{TQA} & \textbf{Avg.} && \textbf{GMS8K} & \textbf{MATH} & \textbf{SVAMP} & \textbf{ARC} & \textbf{CSQA} & \textbf{TQA} & \textbf{Avg.} \\
        \cmidrule(lr){2-8}  \cmidrule(lr){10-16}
        & \multicolumn{7}{c}{OLMo 2 (1B and 13B)} & & \multicolumn{7}{c}{Qwen 3 (1.7B and 14B)} \\
        \midrule
        \multicolumn{16}{l}{\textbf{Building blocks}}\\
        \midrule
        $\vp$      - AR                   & $26.6$ & $26.4$ & $26.3$ & $27.1$ & $27.1$ & $26.9$ & $26.7$ && $26.8$ & $26.5$ & $26.1$ & $26.4$ & $26.6$ & $26.5$ & $26.4$ \\
        $\vq^\ast$ - AR                   & $65.8$ & $65.7$ & $64.9$ & $66.5$ & $66.5$ & $65.9$ & $65.8$ && $39.4$ & $39.0$ & $39.1$ & $39.4$ & $38.8$ & $39.0$ & $39.1$ \\
        
        \midrule
        \multicolumn{16}{l}{\textbf{Nudging \cite{fei2025nudging}}}\\
        \midrule
        
        $\lambda = 0.4$ - AR          & $26.6$ & $26.8$ & $26.7$ & $26.0$ & $24.5$ & $25.5$ & $26.0$ && $26.2$ & $26.1$ & $26.2$ & $26.1$ & $25.9$ & $26.1$ & $26.0$ \\
        $\lambda = 0.4, \gamma = 3$   & $27.2$ & $27.1$ & $26.8$ & $25.6$ & $25.3$ & $25.7$ & $26.2$ && $21.5$ & $21.1$ & $21.5$ & $19.7$ & $18.8$ & $19.4$ & $20.3$ \\
        $\lambda = 0.4, \gamma = 5$   & $29.1$ & $28.6$ & $28.5$ & $25.4$ & $24.8$ & $26.2$ & $27.0$ && $22.7$ & $22.4$ & $22.3$ & $19.2$ & $18.0$ & $18.8$ & $20.5$ \\
        $\lambda = 0.4, \gamma = 7$   & $28.5$ & $27.6$ & $26.9$ & $23.9$ & $22.8$ & $23.8$ & $25.5$ && $22.0$ & $21.5$ & $21.5$ & $17.4$ & $15.9$ & $17.0$ & $19.2$ \\
        
        \midrule
        \multicolumn{16}{l}{\textbf{Our work: dual approximation}}\\
        \midrule
        \multicolumn{4}{l}{$\hookrightarrow$ $\vpi^{<\lambda}$ with $\lambda = f^{-1}(\mu)$}&&&&&&&&&&&&
        \\[0.1cm]
        $\lambda = 0.4$ - AR         & $19.4$ & $19.3$ & $19.4$ & $19.6$ & $19.3$ & $19.5$ & $19.4$ && $16.1$ & $16.0$ & $16.0$ & $16.1$ & $16.1$ & $15.9$ & $16.0$ \\
        $\lambda = 0.4, \gamma = 3$  & $28.8$ & $28.6$ & $28.0$ & $27.7$ & $\underline{28.2}$ & $27.8$ & $28.1$ && $22.1$ & $22.0$ & $22.2$ & $\mathbf{20.6}$ & $\mathbf{19.8}$ & $\underline{20.2}$ & $21.1$ \\
        $\lambda = 0.4, \gamma = 5$  & $\mathbf{31.9}$ & $\underline{31.2}$ & $\underline{30.2}$ & $\mathbf{28.9}$ & $\mathbf{30.3}$ & $\mathbf{29.3}$ & $\mathbf{30.3}$ && $\mathbf{23.6}$ & $\mathbf{23.0}$ & $\mathbf{23.2}$ & $\underline{20.2}$ & $\underline{19.6}$ & $\mathbf{20.3}$ & $\underline{21.6}$ \\
        $\lambda = 0.4, \gamma = 7$  & $\underline{31.7}$ & $\mathbf{31.6}$ & $\mathbf{30.3}$ & $27.4$ & $27.7$ & $\underline{28.4}$ & $\underline{29.5}$ && $23.1$ & $22.6$ & $22.9$ & $19.1$ & $17.8$ & $18.8$ & $20.7$ \\
        
        \midrule
        \multicolumn{16}{l}{\textbf{Our work: primal approximation}}\\
        \midrule
        
        $b = 0.9$ - AR               & $19.3$ & $19.1$ & $19.4$ & $19.3$ & $19.2$ & $19.4$ & $19.2$ && $15.8$ & $16.0$ & $15.8$ & $15.9$ & $15.9$ & $16.1$ & $15.9$ \\
        $b = 0.9, \gamma = 3$        & $28.7$ & $28.1$ & $28.1$ & $26.9$ & $26.8$ & $26.8$ & $27.5$ && $22.0$ & $21.7$ & $22.2$ & $20.0$ & $\mathbf{19.8}$ & $20.0$ & $\mathbf{21.9}$ \\
        $b = 0.9, \gamma = 5$        & $30.6$ & $30.3$ & $29.8$ & $\underline{27.8}$ & $25.5$ & $26.6$ & $28.4$ && $\underline{23.5}$ & $\underline{22.9}$ & $\underline{23.1}$ & $\underline{20.2}$ & $19.1$ & $19.9$ & $21.4$ \\
        $b = 0.9, \gamma = 7$        & $30.1$ & $29.8$ & $28.7$ & $24.6$ & $21.6$ & $22.4$ & $26.2$ && $23.4$ & $22.4$ & $22.5$ & $18.2$ & $17.4$ & $18.2$ & $20.3$ \\
        \bottomrule
    \end{tabular}
    \caption{Model's throughput in tokens per second. ``AR'' stands for auto-regressive, other lines correspond to speculative decoding with a draft size of $\gamma$. Best results on each column are in bold, second best are underlined.}
    \label{tab:throughput}
\end{table*}
\setlength{\tabcolsep}{2.5pt}
\begin{table*}[!ht]
    \small
    \centering
    \begin{tabular}{l ccc ccc a c ccc ccc a}
        \toprule
        & \textbf{GMS8K} & \textbf{MATH} & \textbf{SVAMP} & \textbf{ARC} & \textbf{CSQA} & \textbf{TQA} & \textbf{Avg.} && \textbf{GMS8K} & \textbf{MATH} & \textbf{SVAMP} & \textbf{ARC} & \textbf{CSQA} & \textbf{TQA} & \textbf{Avg.} \\
        \cmidrule(lr){2-8}  \cmidrule(lr){10-16}
        & \multicolumn{7}{c}{OLMo 2 (1B and 13B)} & & \multicolumn{7}{c}{Qwen 3 (1.7B and 14B)} \\
        \midrule
        \multicolumn{16}{l}{\textbf{Nudging \cite{fei2025nudging}}}\\
        \midrule
        
        $\lambda = 0.4, \gamma = 3$ & $67.6$ & $69.2$ & $65.9$ & $59.0$ & $56.5$ & $59.3$ & $62.9$ && $78.4$ & $78.1$ & $\underline{75.3}$ & $65.0$ & $59.5$ & $64.0$ & $70.0$ \\
        $\lambda = 0.4, \gamma = 5$ & $56.5$ & $57.5$ & $53.3$ & $46.7$ & $44.0$ & $48.2$ & $51.0$ && $69.5$ & $69.0$ & $65.9$ & $53.2$ & $47.5$ & $52.2$ & $59.5$ \\
        $\lambda = 0.4, \gamma = 7$ & $48.0$ & $48.6$ & $43.1$ & $38.6$ & $36.6$ & $38.5$ & $42.2$ && $61.7$ & $60.9$ & $59.2$ & $44.4$ & $38.6$ & $43.6$ & $51.4$ \\[0.2cm]
        
        \midrule
        \multicolumn{16}{l}{\textbf{Our work: dual approximation}}\\
        \midrule
        \multicolumn{16}{l}{$\hookrightarrow$ $\vpi^{<\lambda}$ with $\lambda = f^{-1}(\mu)$}
        \\[0.1cm]
        
        $\lambda = 0.4, \gamma = 3$ & $\mathbf{73.7}$ & $\mathbf{75.6}$ & $\mathbf{71.4}$ & $\mathbf{67.7}$ & $\mathbf{69.3}$ & $\mathbf{68.8}$ & $\mathbf{71.0}$ && $\underline{81.6}$ & $\mathbf{81.3}$ & $\mathbf{79.8}$ & $\mathbf{69.9}$ & $\mathbf{66.2}$ & $\mathbf{70.2}$ & $\mathbf{74.8}$ \\
        $\lambda = 0.4, \gamma = 5$ & $64.0$ & $65.9$ & $59.3$ & $55.5$ & $60.0$ & $57.3$ & $60.3$ && $73.9$ & $72.5$ & $71.6$ & $59.1$ & $54.7$ & $59.5$ & $65.2$ \\
        $\lambda = 0.4, \gamma = 7$ & $55.4$ & $57.9$ & $52.3$ & $47.6$ & $48.3$ & $49.8$ & $51.8$ && $67.0$ & $65.3$ & $64.0$ & $50.1$ & $46.0$ & $50.3$ & $57.1$ \\
        
        \midrule
        \multicolumn{16}{l}{\textbf{Our work: primal approximation}}\\
        \midrule
        $b = 0.9, \gamma = 3$       & $\underline{73.5}$ & $\underline{74.6}$ & $\underline{68.8}$ & $\underline{65.5}$ & $\underline{64.9}$ & $\underline{64.4}$ & $\underline{68.6}$ && $\mathbf{81.7}$ & $\underline{80.9}$ & $\mathbf{79.8}$ & $\underline{69.8}$ & $\underline{66.0}$ & $\underline{69.4}$ & $\underline{74.6}$ \\
        $b = 0.9, \gamma = 5$       & $62.0$ & $63.1$ & $57.2$ & $53.0$ & $46.7$ & $49.9$ & $55.3$ && $74.0$ & $72.2$ & $71.3$ & $58.1$ & $53.1$ & $57.6$ & $64.3$ \\
        $b = 0.9, \gamma = 7$       & $53.5$ & $54.5$ & $48.0$ & $41.2$ & $33.8$ & $36.4$ & $44.5$ && $66.8$ & $64.3$ & $64.2$ & $48.5$ & $44.1$ & $48.1$ & $56.0$ \\[0.2cm]
        \bottomrule
    \end{tabular}
    \caption{Speculative generation empirical acceptance rate. Best results on each column are in bold, second best are underlined.}
    \label{tab:acceptance_rate}
\end{table*}
\section{Generation Speed Results}
\label{app:generation_speed}

Table~\ref{tab:throughput} shows the throughput for all the tested models, both for autoregressive generation and speculative decoding.
Table~\ref{tab:acceptance_rate} shows the exact acceptance rates.

\end{document}